\documentclass{article} 
\usepackage{iclr2016_conference,times}
\usepackage[hyphens]{url}
\usepackage{hyperref}
\usepackage{times}
\usepackage{latexsym}
\usepackage{amsmath,amssymb}
\usepackage{multirow}
\usepackage{color}
\usepackage{graphicx}
\usepackage{bbm}
\usepackage{xspace}
\usepackage{wasysym}
\usepackage{latexsym}
\usepackage{graphicx}
\usepackage{algorithmic}
\usepackage{float}
\usepackage{mathtools}
\usepackage[titletoc,title]{appendix}
\usepackage{pgfplots}
\usepackage{placeins}
\usepackage{tabularx}
\newcolumntype{L}[1]{>{\raggedright\arraybackslash}p{#1}}
\newcolumntype{C}[1]{>{\centering\arraybackslash}p{#1}}
\newcolumntype{R}[1]{>{\raggedleft\arraybackslash}p{#1}}
 
\mathtoolsset{showonlyrefs}  
\definecolor{mbgreen}{RGB}{34,189,64}

\newcommand{\igate}{i}
\newcommand{\fgate}{f}
\newcommand{\ogate}{o}
\newcommand{\state}{c}
\newcommand{\wtmat}[2]{W_{#1 #2}}

\newcommand{\annoppdb}{Annotated-PPDB\xspace}
\newcommand{\paragramsl}{\textsc{paragram-sl999}\xspace}
\newcommand{\paragramphrase}{\textsc{paragram-phrase}\xspace}
\newcommand{\rev}[1]{\textcolor{black}{#1}}

\newcommand\norm[1]{\left\lVert#1\right\rVert}
\DeclareMathOperator*{\argmax}{argmax}

\title{Towards Universal\\Paraphrastic Sentence Embeddings}
\author{
   John Wieting
\ \ \ \  Mohit Bansal
\ \ \ \  Kevin Gimpel
\ \ \ \  Karen Livescu
\\
Toyota Technological Institute at Chicago, Chicago, IL, 60637, USA\\
\tt{\{jwieting,mbansal,kgimpel,klivescu\}@ttic.edu}
}

\iclrfinalcopy 

\begin{document}

\maketitle

\begin{abstract}
We consider the problem of 
learning general-purpose, paraphrastic sentence embeddings based on supervision from the
Paraphrase Database~\citep{GanitkevitchDC13}.
We compare six compositional architectures, evaluating them on annotated
textual similarity datasets drawn both from the same distribution as
the training data 
and 
from a wide range of 
other domains. 
We find that the most complex architectures, such as long short-term memory (LSTM) recurrent neural networks, perform best on the in-domain data. 
However, 
in 
out-of-domain scenarios, simple architectures 
such as word averaging vastly outperform LSTMs. Our simplest averaging model is even competitive with systems tuned for the particular tasks while also being extremely efficient and easy to use.

In order to better understand how these architectures compare, we conduct further experiments on three supervised NLP tasks: sentence similarity, entailment, and sentiment classification. 
We again find that the word averaging models perform well for sentence similarity and entailment, outperforming LSTMs. 
However, on sentiment classification, we find that the LSTM performs
very strongly\----even recording new state-of-the-art performance on
the Stanford Sentiment Treebank.

We then demonstrate how to combine our pretrained 
sentence embeddings 
with these supervised tasks, using them both as a prior and as a black box 
feature extractor. 
This leads to performance rivaling the state of the art on the SICK similarity and entailment tasks.
We release 
all of our resources 
to the research community\footnote{Trained models and code for training and evaluation are available at \url{http://ttic.uchicago.edu/~wieting}.} with the hope that they can serve as the new baseline for further work on universal 
sentence embeddings.

\end{abstract}

\section{Introduction}
Word embeddings have become ubiquitous in natural language processing (NLP). Several researchers have developed and shared word embeddings trained on large datasets~\citep{Collobert:2011:NLP,mikolov2013distributed,pennington2014glove}, and these have been used effectively for many downstream tasks~\citep{Turian10wordrepresentations,SocherEtAl2011:PoolRAE,kim-14,bansal2014tailoring,tai2015improved}.  There has also been recent work on creating representations for word sequences such as phrases or sentences. Many functional architectures have been proposed to model compositionality in such sequences, ranging from those based on simple operations like addition~\citep{Mitchell:Lapata:2010,TACL586,iyyer-EtAl:2015:ACL-IJCNLP} to those based on richly-structured functions like recursive neural networks~\citep{SocherEtAl2011:PoolRAE}, convolutional neural networks~\citep{kalchbrenner-grefenstette-blunsom:2014:P14-1}, and recurrent neural networks using long short-term memory (LSTM)~\citep{tai2015improved}.
However, there is little work on learning sentence representations
that can be used across domains with the same ease and effectiveness
as word embeddings. In this paper, we explore 
compositional models that can encode arbitrary word sequences into a vector with the property that sequences with similar meaning have high cosine similarity, and that can, importantly, also transfer easily across domains. We consider six compositional architectures based on neural networks and train them on noisy phrase pairs from the Paraphrase Database~(PPDB; Ganitkevitch et al., 2013).

We consider models spanning the range of complexity from word averaging to LSTMs. 
With the 
simplest word averaging model, there are no additional compositional parameters. 
The only parameters 
are the word vectors themselves,
which are learned 
to produce effective sequence embeddings 
when averaging is performed over the sequence. 
We add
complexity 
by adding layers, 
leading to variants of deep averaging networks~\citep{iyyer-EtAl:2015:ACL-IJCNLP}. We next consider several recurrent network variants, culminating 
in LSTMs because they have been found to be effective for many types of sequential data~\citep{NIPS2007_3213,graves2013speech,greff2015lstm}, including text~\citep{sutskever2014sequence,vinyals2014grammar,DBLP:conf/icml/XuBKCCSZB15,nips15_hermann,ling-EtAl:2015:EMNLP2,wen-EtAl:2015:EMNLP}. 

To evaluate our models, we consider two 
tasks drawn from the same 
distribution as the training data, as well as 22 SemEval textual similarity datasets from a variety of domains (such as news, tweets, web forums, and image and video captions). 
Interestingly, we find that the LSTM performs well on the in-domain task, but performs much worse on the out-of-domain tasks.  
We discover surprisingly strong performance for the models based on
word averaging, which perform well on both the in-domain and
out-of-domain tasks, beating the best LSTM model by 16.5 Pearson's $r$ on
average.
Moreover, we find that learning word embeddings in the context of
vector averaging performs much better than simply averaging
pretrained, state-of-the-art word embeddings. Our average Pearson's
$r$ over all 22 SemEval datasets is 17.1 points higher than averaging
GloVe vectors\footnote{We used the publicly available 300-dimensional
vectors that were trained on the 840 billion token Common Crawl
corpus, available at \url{http://nlp.stanford.edu/projects/glove/}.}
and 12.8 points higher than averaging \paragramsl
vectors.\footnote{These are 300-dimensional vectors from
\cite{wieting2015ppdb} and are available at
\url{http://ttic.uchicago.edu/~wieting}. They give human-level
performance on two commonly used word similarity datasets, WordSim353 \citep{finkelstein2001placing} and Simlex-999 \citep{HillRK14}.} 

Our final sentence embeddings\footnote{Denoted \paragramphrase-XXL and discussed
in Section~\ref{sec:XXL}.} place in the top 25\% of all submitted
systems in every SemEval STS task from 2012 through 2015, being best or tied for best 
on 4 of the datasets.\footnote{As measured by the average Pearson's $r$ over all datasets in each task; \rev{ see Table~\ref{fig:stats}}.} 
This is surprising because the submitted systems were designed for those particular tasks, with access to training and tuning data specifically developed for each task.  

While the above experiments focus on transfer, we also consider the fully supervised setting (Table~\ref{fig:Supervised}). 
We compare the same suite of compositional architectures for three supervised NLP tasks: sentence similarity and textual entailment using the 2014 SemEval SICK dataset~\citep{marelli2014semeval}, and sentiment classification using the Stanford Sentiment Treebank~\citep{socher-13}. 
We again find strong performance for the word averaging models \rev{for both similarity and entailment}, outperforming the LSTM. 
\rev{However, for sentiment classification, we see a different trend. The LSTM now performs best, achieving 89.2\% on the coarse-grained sentiment classification task.  
This result, to our knowledge, is the new state of the art on this task.} 

We then demonstrate how to combine our PPDB-trained 
sentence embedding models with supervised NLP tasks. 
We first use our model as a prior, yielding performance 
on the similarity and entailment tasks that rivals the state of the art. 
We also use our sentence embeddings as an effective 
black box
feature extractor for downstream tasks, comparing favorably to recent work~\citep{kiros2015skip}. 

We release our strongest sentence embedding model, which we call \paragramphrase XXL, to the research community.\footnote{Available at \url{http://ttic.uchicago.edu/~wieting}.} Since it consists merely of a new set of word embeddings, it is extremely efficient and easy to use for downstream applications. Our hope is that this model can provide a new simple and strong baseline in the quest for universal sentence embeddings. 

\section{Related Work}
Researchers have developed many ways to embed word sequences for NLP. They mostly focus on the question of compositionality: given vectors for words, how should we create a vector for a word sequence? \cite{mitchell2008vector,Mitchell:Lapata:2010} considered bigram compositionality, comparing many functions for composing two word vectors into a single vector to represent their bigram. Follow-up work by \cite{Blacoe2012} found again that simple operations such as vector addition performed strongly. Many other compositional architectures have been proposed. 
\rev{Some have been based on distributional semantics~\citep{baroni2014frege,paperno-pham-baroni:2014:P14-1,polajnar-rimell-clark:2015:LSDSem,DBLP:journals/corr/TianOI15}, while the current trend is toward development of neural network architectures.}
These include neural bag-of-words models~\citep{kalchbrenner-grefenstette-blunsom:2014:P14-1}, \rev{deep averaging networks (DANs)}~\citep{iyyer-EtAl:2015:ACL-IJCNLP}, feature-weighted averaging~\citep{TACL586}, recursive neural networks based on parse structure~\citep{SocherEtAl2011:PoolRAE,socher-EtAl:2012:EMNLP-CoNLL,socher-13,irsoy-drsv,wieting2015ppdb}, recursive networks based on non-syntactic hierarchical structure~\citep{zhao2015self,chen-EtAl:2015:EMNLP1},  convolutional neural networks~\citep{kalchbrenner-grefenstette-blunsom:2014:P14-1,kim-14,NIPS2014_5550,yin-schutze:2015:NAACL-HLT1,he-gimpel-lin:2015:EMNLP}, and recurrent neural networks using long short-term memory~\citep{tai2015improved,ling-EtAl:2015:EMNLP2,liu-EtAl:2015:EMNLP2}. 
In this paper, we compare six architectures: word averaging, word averaging followed by a single linear projection, DANs, and three variants of recurrent neural networks, including LSTMs.\footnote{In prior work, we experimented with recursive neural networks on binarized parses of the PPDB~\citep{wieting2015ppdb}, but we found that many of the phrases in PPDB are not sentences or even constituents, causing the parser to have unexpected behavior.}  

Most of the work mentioned above learns compositional models in the context of \emph{supervised} learning. That is, a training set is provided with annotations and the composition function is learned for the purposes of optimizing an objective function based on those annotations. The models are then evaluated on a test set drawn from the same distribution as the training set. 

In this paper, in contrast, we are primarily interested in creating
general purpose, domain independent embeddings for word
sequences. There have been 
research efforts also targeting this goal. 
One approach is to train an autoencoder in an attempt to learn the latent structure of the sequence, whether it be a sentence with a parse tree~\citep{SocherEtAl2011:PoolRAE}, or a longer sequence such as a paragraph or document~\citep{li-luong-jurafsky:2015:ACL-IJCNLP}. Other recently proposed methods, including paragraph vectors~\citep{le2014distributed} and skip-thought vectors~\citep{kiros2015skip}, learn sequence representations that are predictive of words inside the sequence or in neighboring sequences. 
These methods produce generic representations that can be used to provide features for text classification or sentence similarity tasks. 
While skip-thought vectors capture similarity in terms of discourse context, in this paper we are interested in 
capturing paraphrastic similarity, i.e., whether two sentences have the same meaning. 

\rev{Our learning formulation draws from a large body of related work on learning input representations in order to maximize similarity in the learned space~\citep{weston2010large,yih-EtAl:2011:CoNLL,huang2013learning,hermann-blunsom:2014:P14-1,SocherKLMN14,faruqui-dyer:2014:EACL,bordes2014open,D14-1067,lu-EtAl:2015:NAACL-HLT}, including our prior work~\citep{wieting2015ppdb}. 
We focus our exploration here on modeling and keep the learning methodology mostly fixed, though we do include certain choices about the learning procedure in our hyperparameter tuning space for each model.}

\section{Models and Training} \label{sec:models}
Our goal is to embed sequences into a low-dimensional space such that cosine similarity in the space corresponds to the strength of the paraphrase relationship between the sequences. We experimented with six models of increasing complexity. The simplest model embeds a word sequence $x=\langle x_1,x_2,...,x_n\rangle$ by averaging the vectors of its tokens. 
The only parameters learned by this model are the word embedding matrix $W_w$:
\begin{equation}
g_{\paragramphrase}(x) = \frac{1}{n}\sum_i^n W_w^{x_i}
\end{equation}
\noindent where $W_w^{x_i}$ is the word embedding for word $x_i$. 
We call the learned embeddings \paragramphrase embeddings. 

In our second model, we learn a projection in addition to the word embeddings:
\begin{equation}
g_{\mathrm{proj}}(x) = W_p\left(\frac{1}{n}\sum_i^n W_w^{x_i}\right)+b
\end{equation}
\noindent where 
$W_p$ is the projection matrix and $b$ is a bias vector. 

Our third model is the deep averaging network (DAN) of \cite{iyyer-EtAl:2015:ACL-IJCNLP}. This is a generalization of the above models that typically uses 
multiple layers as well as nonlinear activation functions. In our experiments below, we tune over the number of layers and choice of activation function. 

Our fourth model is a standard recurrent network (RNN) with randomly initialized weight matrices and nonlinear activations: 
\begin{align}
&h_t = f(W_{x}W_w^{x_t} + W_{h}h_{t-1} + b)\\
&g_{\mathrm{RNN}}(x) = h_{-1}
\end{align}
\noindent where $f$ is the activation function (either $\tanh$ or rectified linear unit; the choice is tuned), $W_{x}$ and $W_{h}$ are parameter matrices, $b$ is a bias vector, and $h_{-1}$ refers to the hidden vector of the last token.

Our fifth model is a special RNN which we call an {\it identity-RNN}. In the identity-RNN, the weight matrices are initialized to identity, the bias is initialized to zero, and the activation is the identity function. We divide the final output vector of the identity-RNN by the number of tokens in the sequence. Thus, before any updates to the parameters, the identity-RNN simply averages the word embeddings. We also regularize the identity-RNN parameters to their initial values. The idea is that, with high regularization, the identity-RNN is simply averaging word embeddings. However, it is a richer architecture and can take into account word order and hopefully improve upon the averaging baseline. 

Our sixth and final model is the most expressive. We use long short-term memory (LSTM)~\citep{hochreiter1997long}, a recurrent neural network (RNN) architecture designed to model sequences with long-distance dependencies. 
LSTMs have recently been shown to produce state-of-the-art results in a variety of sequence processing tasks~\citep{chen-EtAl:2015:EMNLP2,filippova-EtAl:2015:EMNLP,xu-EtAl:2015:EMNLP2,belinkov-glass:2015:EMNLP,wang-nyberg:2015:ACL-IJCNLP}. We use the version from~\cite{gers2003learning} which has the following equations:
\begin{align}
&\igate_t = \sigma\left(\wtmat{x}{\igate} W_w^{x_t} + \wtmat{h}{\igate} h_{t-1} + \wtmat{\state}{\igate} \state_{t-1}  + b_\igate\right)\\
&\fgate_t = \sigma\left(\wtmat{x}{\fgate} W_w^{x_t} + \wtmat{h}{\fgate} h_{t-1} + \wtmat{\state}{\fgate} \state_{t-1} + b_\fgate \right)\\
&\state_t = \fgate_t \state_{t-1} + \igate_t \tanh \left(\wtmat{x}{\state} W_w^{x_t} + \wtmat{h}{\state} h_{t-1} + b_\state\right)\\
&\ogate_t = \sigma\left(\wtmat{x}{\ogate} W_w^{x_t} + \wtmat{h}{\ogate} h_{t-1} + \wtmat{\state}{\ogate} \state_{t} + b_\ogate\right)\\
&h_t = \ogate_t \tanh(\state_t)\\
&g_{\mathrm{LSTM}}(x) = h_{-1}
\end{align}
\noindent where $\sigma$ is the logistic sigmoid function.
We found that the choice of whether or not to include the output gate had a significant impact on performance, so we used two versions of the LSTM model, one with the output gate and one without. 
For all models, we learn the word embeddings themselves, denoting the trainable word embedding parameters by $W_w$. We denote all other trainable parameters by $W_c$ (``compositional parameters''), 
though the \paragramphrase model has no compositional parameters. 
We initialize $W_w$ using some embeddings pretrained from large corpora. 

\subsection{Training}

We mostly follow the approach of \cite{wieting2015ppdb}. 
The training data consists of (possibly noisy) pairs taken directly from the original Paraphrase Database (PPDB) and we optimize a margin-based loss. 

Our training data consists of a set $X$ of phrase pairs $\langle x_1, x_2\rangle$, where $x_1$ and $x_2$ are assumed to be paraphrases. The objective function follows:
\begin{align}
\underset{W_c,W_w}{\text{min}} \frac{1}{|X|}\Bigg(\sum_{\langle x_1,x_2\rangle \in X} 
&\max(0,\delta - \cos(g(x_1), g(x_2)) + \cos(g(x_1), g(t_1)))  \nonumber\\
+ &\max(0,\delta - \cos(g(x_1),g(x_2)) + \cos(g(x_2), g(t_2)))\bigg) \nonumber\\
+ &\lambda_c\norm{W_c}^2  + \lambda_{w}\norm{W_{w_{\mathit{initial}}} - W_w}^2\label{eq:phrase}
\end{align} 
\noindent where $g$ is the embedding function in use (e.g., $g_{\mathrm{LSTM}}$), $\delta$ is the margin, $\lambda_{c}$ and $\lambda_{w}$ are regularization parameters, $W_{w_{\mathit{initial}}}$ is the initial word embedding matrix, and $t_1$ and $t_2$ are carefully-selected negative examples taken from a mini-batch during optimization.  
The intuition is that we want the two phrases to be more similar to each other ($\cos(g(x_1), g(x_2))$) than either is to their respective negative examples $t_1$ and $t_2$, by a margin of at least $\delta$. 

\subsubsection{Selecting Negative Examples}
\label{sec:sample}
To select $t_1$ and $t_2$ in Eq.~\refeq{eq:phrase}, we tune the choice between two approaches. The first, MAX, simply chooses the most similar phrase in some set of phrases (other than those in the given phrase pair). For simplicity and to reduce the number
of tunable parameters, we use the mini-batch for this set, but it could be a separate set. Formally, MAX corresponds to choosing $t_1$ for a given $\langle x_1, x_2\rangle$ as follows:
\begin{equation}
t_1 = \argmax_{t : \langle t, \cdot\rangle \in X_b \setminus \{\langle x_1, x_2\rangle\}} \cos(g(x_1), g(t))
\end{equation}
\noindent where $X_b\subseteq X$ is the current mini-batch. 
That is, we want to choose a negative example $t_i$ that is similar to $x_i$ according to the current model parameters. 
The downside of this approach is that we may occasionally choose a phrase $t_i$ that is actually a true paraphrase of $x_i$. 

The second strategy selects negative examples using MAX with probability 0.5 and selects them randomly from the mini-batch otherwise. We call this sampling strategy MIX. We tune over the strategy in our experiments.

\section{Experiments}
\subsection{Data}
We experiment on 24 textual similarity datasets, covering many domains, including all datasets from every SemEval semantic textual similarity (STS) task (2012-2015). We also evaluate on the SemEval 2015 Twitter task \citep{xu2015semeval} and the SemEval 2014 Semantic Relatedness task \citep{marelli2014semeval}, as well as two tasks that use PPDB data \citep{wieting2015ppdb,PavlickEtAl-2015:ACL:PPDB2.0}.

The first STS task was held in 2012 and these tasks have been held every year since. Given two sentences, the objective of the task is to predict how similar they are on a 0-5 scale, where 0 indicates the sentences are on different topics and 5 indicates that they are completely equivalent. Each STS task consists of 4-6 different datasets and the tasks cover a wide variety of domains which we have categorized below. Most submissions for these tasks use supervised models that are trained and tuned on either provided training data or 
similar datasets from older tasks. Details on the number of teams and submissions for each task and the performance of the submitted systems for each dataset are included in Table~\ref{fig:STS_tasks} and Table~\ref{fig:STS results} respectively. For more details on these tasks please refer to the relevant publications for the 2012~\citep{agirre2012semeval}, 2013~\citep{diab2013eneko}, 2014~\citep{agirre2014semeval}, and 2015~\citep{agirre2015semeval} tasks. 

\begin{table}[h]
\setlength{\tabcolsep}{4pt}
\small
\centering
\begin{tabular} {| l | c | c |} \hline
Dataset & No. of teams & No. of submissions\\
\hline
2012 STS &  35 & 88 \\
2013 STS &  34 & 89 \\
2014 STS & 15 & 38 \\
2015 STS & 29 & 74 \\
2014 SICK & 17 & 66 \\
2015 Twitter & 19 & 26 \\
\hline
\end{tabular}
\caption{\label{fig:STS_tasks}
Details on numbers of teams and submissions in the STS tasks used for evaluation. 
\vspace{-0.3cm}}
\end{table}

Below are the textual domains contained in the STS tasks:\\
{\bf News}: Newswire was used in the 2012 task (MSRpar) and the 2013 and 2014 tasks (deft news). \\
{\bf Image and Video Descriptions}: Image descriptions generated via crowdsourcing were used in the 2013 and 2014 tasks (images). Video descriptions were used in the 2012 task (MSRvid).\\
{\bf Glosses}: Glosses from WordNet, OntoNotes, and FrameNet were used in the 2012, 2013, and 2014 tasks (OnWN and FNWN).\\
{\bf MT evaluation}: The output of machine translation systems with their reference translations was used in the 2012 task (SMT-eur and SMT-news) and the 2013 task (SMT).\\
{\bf Headlines}: Headlines of news articles were used in the 2013, 2014, and 2015 tasks (headline).\\
{\bf Web Forum}:  Forum posts were used in the 2014 task (deft forum).  \\
{\bf Twitter}: Pairs containing a tweet related to a news headline and a sentence pertaining to the same news headline. This dataset was used in the 2014 task (tweet news). \\
{\bf Belief}: Text from the Deft Committed Belief Annotation (LDC2014E55) was used in the 2015 task (belief). \\
{\bf Questions and Answers}: Paired answers to the same question from StackExchange (answers-forums) and the BEETLE corpus \citep{dzikovska2010beetle} (answers-students) were used in 2015.  

For tuning, we use two datasets that contain PPDB phrase pairs scored by human annotators on the strength of their paraphrase relationship. One is a large sample of 26,456 annotated phrase pairs developed by \cite{PavlickEtAl-2015:ACL:PPDB2.0}. The second, called \annoppdb, was developed in our prior work~\citep{wieting2015ppdb} and is a small set of 1,000 annotated phrase pairs that were filtered to focus on challenging paraphrase phenomena. 

\subsection{Transfer Learning}
\subsubsection{Experimental Settings} \label{sec:tuning}

As training data, we used the XL section\footnote{PPDB comes in different sizes (S, M, L, XL, XXL, and XXXL), where each larger size subsumes all smaller ones. The phrases are sorted by a confidence measure and so the smaller sets contain higher precision paraphrases.} of PPDB which contains 3,033,753 unique phrase pairs. 
However, for hyperparameter tuning 
we only used 100k examples sampled from PPDB XXL and trained for 5 epochs. 
Then after finding the hyperparameters that maximize Spearman's $\rho$ on the Pavlick et al.~PPDB task, 
we trained on the entire XL section of PPDB for 10 epochs. We used \paragramsl embeddings to initialize the word embedding matrix ($W_w$) for all models.

We chose the Pavlick et al.\nocite{PavlickEtAl-2015:ACL:PPDB2.0}~task for tuning because we wanted our entire procedure to only make use of PPDB and use no other resources. In particular, we did not want to use any STS tasks for training or hyperparameter tuning. 
We chose the Pavlick et al.~dataset over \annoppdb due to its larger size. 
But in practice the datasets are very similar and tuning on either 
produces similar results.

To learn model parameters for all experiments in this section, we minimize Eq.~\ref{eq:phrase}. 
Our models have the following tunable hyperparameters:\footnote{For $\lambda_c$ we searched over $\{10^{-3},10^{-4},10^{-5},10^{-6}\}$, for $b$ we searched over $\{25, 50, 100\}$, for $\lambda_w$ we searched over $\{10^{-5}, 10^{-6}, 10^{-7}, 10^{-8}\}$ as well as the setting in which we do not update $W_w$, and for $\delta$ we searched over $\{0.4, 0.6, 0.8\}$.} $\lambda_c$, the L$_2$ regularizer on the compositional parameters $W_c$ (not applicable for the word averaging model), the pool of phrases used to obtain negative examples (coupled with mini-batch size $B$, to reduce the number of tunable hyperparameters), 
$\lambda_w$, the regularizer on the word embeddings, and $\delta$, the margin. We also tune over optimization method (either AdaGrad~\citep{Duchi} or Adam~\citep{kingma2014adam}), learning rate (from $\{0.05,0.005,0.0005\}$), whether to clip the gradients with threshold 1~\citep{pascanu2012difficulty}, 
and whether to use MIX or MAX sampling. For the classic RNN, we further tuned whether to use $\tanh$ or rectified linear unit activation functions; for the identity-RNN, we tuned $\lambda_c$ over $\{1000,100,10,1\}$ because we wanted higher regularization on the composition parameters; for the DANs we tuned over activation function ($\tanh$ or rectified linear unit) and the number of layers (either 1 or 2); for the LSTMs we tuned on whether to include an output gate.  We fix the output dimensionalities of all models that require doing so 
to the dimensionality of our word embeddings (300).

\subsubsection{Results} \label{sec:results}

The results on all STS tasks as well as the SICK and Twitter tasks are shown in Table~\ref{fig:STS results}. We include results on the PPDB tasks in Table~\ref{fig:PPDB results}. 
In Table~\ref{fig:STS results}, we first show the median, $75^\textrm{th}$ percentile, and highest score from the official task rankings. We then report the performance of our seven models: \paragramphrase (PP), identity-RNN (iRNN), projection (proj.), deep-averaging network (DAN), recurrent neural network (RNN), LSTM with output gate (o.g.), and LSTM without output gate (no o.g.). We compare to three baselines: skip-thought vectors\footnote{Note that we pre-processed the training data with the tokenizer from Stanford CoreNLP \citep{manning-EtAl:2014:P14-5} rather than the included NLTK \citep{bird2009natural} tokenizer. We found that doing so significantly improves the performance of the skip-thought vectors.}~\citep{kiros2015skip}, denoted ``ST'', 
averaged GloVe\footnote{We used the publicly available 300-dimensional vectors that were trained on the 840 billion token Common Crawl corpus, available at \url{http://nlp.stanford.edu/projects/glove/}.} vectors~\citep{pennington2014glove}, 
and averaged \paragramsl vectors \citep{wieting2015ppdb}, denoted ``PSL''. 
Note that the GloVe vectors were used to initialize the \paragramsl vectors which were, in turn, used to initialize our \paragramphrase embeddings. We compare to skip-thought vectors because trained models are publicly available and they show impressive performance when used as features on several tasks including textual similarity.

\begin{table}[h]
\setlength{\tabcolsep}{4pt}
\scriptsize
\centering
\begin{tabular} { | l || c | c | c || c | c | c | c | c | C{0.9cm} | C{0.6cm} | c | c | c |} \hline
Dataset & 50\% & 75\% & Max &  PP & proj. & DAN & RNN & iRNN  & LSTM (no o.g.) & LSTM (o.g.) & ST & GloVe & PSL\\
\hline
MSRpar & 51.5 & 57.6 & 73.4 & 42.6 & 43.7 & 40.3 & 18.6 & 43.4 & 16.1 & 9.3 & 16.8 & \bf 47.7 & 41.6\\
MSRvid & 75.5 & 80.3 & 88.0 & \bf 74.5 & 74.0 & 70.0 & 66.5 & 73.4 & 71.3 & 71.3 & 41.7 & 63.9 & 60.0\\
SMT-eur & 44.4 & 48.1 & 56.7 & 47.3 & \bf 49.4 & 43.8 & 40.9 & 47.1 & 41.8 & 44.3 & 35.2 & 46.0 & 42.4\\
OnWN & 608 & 65.9 & 72.7 & \bf 70.6 & 70.1 & 65.9 & 63.1 & 70.1 & 65.2 & 56.4 & 29.7 & 55.1 & 63.0\\
SMT-news & 40.1 & 45.4 & 60.9 & 58.4 & \bf 62.8 & 60.0 & 51.3 & 58.1 & 60.8 & 51.0 & 30.8 & 49.6 & 57.0\\
\hline
STS 2012 Average & 54.5 & 59.5 & 70.3 & 58.7 & \bf 60.0 & 56.0 & 48.1 & 58.4 & 51.0 & 46.4 & 30.8 & 52.5 & 52.8\\
\hline
headline & 64.0 & 68.3 & 78.4 & 72.4 & 72.6 & 71.2 & 59.5 & \bf 72.8 & 57.4 & 48.5 & 34.6 & 63.8 & 68.8\\
OnWN & 52.8 & 64.8 & 84.3 & 67.7 & 68.0 & 64.1 & 54.6 & \bf 69.4 & 68.5 & 50.4 & 10.0 & 49.0 & 48.0\\
FNWN & 32.7 & 38.1 & 58.2 & 43.9 & \bf 46.8 & 43.1 & 30.9 & 45.3 & 24.7 & 38.4 & 30.4 & 34.2 & 37.9\\
SMT & 31.8 & 34.6 & 40.4 & 39.2 & \bf 39.8 & 38.3 & 33.8 & 39.4 & 30.1 & 28.8 & 24.3 & 22.3 & 31.0\\
\hline
STS 2013 Average & 45.3 & 51.4 & 65.3 & 55.8 & \bf 56.8 & 54.2 & 44.7 & 56.7 & 45.2 & 41.5 & 24.8 & 42.3 & 46.4\\
\hline
deft forum & 36.6 & 46.8 & 53.1 & 48.7 & \bf 51.1 & 49.0 & 41.5 & 49.0 & 44.2 & 46.1 & 12.9 & 27.1 & 37.2\\
deft news & 66.2 & 74.0 & 78.5 & \bf 73.1 & 72.2 & 71.7 & 53.7 & 72.4 & 52.8 & 39.1 & 23.5 & 68.0 & 67.0\\
headline & 67.1 & 75.4 & 78.4 & 69.7 & \bf 70.8 & 69.2 & 57.5 & 70.2 & 57.5 & 50.9 & 37.8 & 59.5 & 65.3\\
images & 75.6 & 79.0 & 83.4 & \bf 78.5 & 78.1 & 76.9 & 67.6 & 78.2 & 68.5 & 62.9 & 51.2 & 61.0 & 62.0\\
OnWN & 78.0 & 81.1 & 87.5 & 78.8 & \bf 79.5 & 75.7 & 67.7 & 78.8 & 76.9 & 61.7 & 23.3 & 58.4 & 61.1\\
tweet news & 64.7 & 72.2 & 79.2 & 76.4 & 75.8 & 74.2 & 58.0 & \bf 76.9 & 58.7 & 48.2 & 39.9 & 51.2 & 64.7\\
\hline
STS 2014 Average & 64.7 & 71.4 & 76.7 & 70.9 & \bf 71.3 & 69.5 & 57.7 & 70.9 & 59.8 & 51.5 & 31.4 & 54.2 & 59.5\\
\hline
answers-forums & 61.3 & 68.2 & 73.9 & \bf 68.3 & 65.1 & 62.6 & 32.8 & 67.4 & 51.9 & 50.7 & 36.1 & 30.5 & 38.8\\
answers-students & 67.6 & 73.6 & 78.8 & \bf 78.2 & 77.8 & 78.1 & 64.7 & 78.2 & 71.5 & 55.7 & 33.0 & 63.0 & 69.2\\
belief & 67.7 & 72.2 & 77.2 & \bf 76.2 & 75.4 & 72.0 & 51.9 & 75.9 & 61.7 & 52.6 & 24.6 & 40.5 & 53.2\\
headline & 74.2 & 80.8 & 84.2 & 74.8 & \bf 75.2 & 73.5 & 65.3 & 75.1 & 64.0 & 56.6 & 43.6 & 61.8 & 69.0\\
images & 80.4 & 84.3 & 87.1 & \bf 81.4 & 80.3 & 77.5 & 71.4 & 81.1 & 70.4 & 64.2 & 17.7 & 67.5 & 69.9\\
\hline
STS 2015 Average & 70.2 & 75.8 & 80.2 & \bf 75.8 & 74.8 & 72.7 & 57.2 & 75.6 & 63.9 & 56.0 & 31.0 & 52.7 & 60.0\\
\hline
2014 SICK & 71.4 & 79.9 & 82.8 & 71.6 & \bf 71.6 & 70.7 & 61.2 & 71.2 & 63.9 & 59.0 & 49.8 & 65.9 & 66.4\\
\hline
2015 Twitter & 49.9 & 52.5 & 61.9 & 52.9 & 52.8 & \bf 53.7 & 45.1 & 52.9 & 47.6 & 36.1 & 24.7 & 30.3 & 36.3\\
\hline
\end{tabular}
\caption{\label{fig:STS results}
Results on SemEval textual similarity datasets (Pearson's $r \times 100$). The highest score in each row is in boldface (omitting the official task score columns). 
\vspace{-0.3cm}
}
\end{table}

The results in Table~\ref{fig:STS results} show strong performance of our two simplest models: the \paragramphrase embeddings (PP) and our projection model (proj.). They outperform the other models on all but 5 of the 22 datasets. 
The iRNN model has the next best performance, while the LSTM models lag behind. These results stand in marked contrast to those in Table~\ref{fig:PPDB results}, which shows very similar performance across models on the in-domain PPDB tasks, with the LSTM models slightly outperforming the others. 
For the LSTM models, it is also interesting to note that removing the output gate results in stronger performance on the textual similarity tasks. Removing the output gate improves performance 
on 18 of the 22 datasets. The LSTM without output gate also performs reasonably well compared to our strong \paragramsl addition baseline, beating it on 12 of the 22 datasets.  

\begin{table}
\footnotesize
\centering
  \begin{tabular}{| l || C{1.9cm} | C{1.9cm} | C{2.3cm} |}
    \hline
    Model & Pavlick et al.\nocite{PavlickEtAl-2015:ACL:PPDB2.0} (oracle) & Pavlick et al.\nocite{PavlickEtAl-2015:ACL:PPDB2.0} (test)  & \annoppdb\nocite{wieting2015ppdb} (test)\\
    \hline
    \paragramphrase & 60.3 & 60.0 & 53.5\\
    projection & 61.0 & 58.4 & 52.8\\
    DAN & 60.9 & 60.1 & 52.3\\
    RNN &60.5 & 60.3 & 51.8\\
    iRNN & 60.3 & 60.0 & \bf 53.9\\
    LSTM (no o.g.) & \bf 61.6 & \bf 61.3 & 53.4\\
    LSTM (o.g.) & 61.5 & 60.9 & 52.9\\
    skip-thought & 39.3 & 39.3 & 31.9\\
    GloVe & 44.8 & 44.8 & 25.3\\
    \paragramsl & 55.3 & 55.3 & 40.4\\
    \hline
   \end{tabular}
  \caption{\label{fig:PPDB results}
Results on the PPDB tasks (Spearman's $\rho \times 100$). For the task in \cite{PavlickEtAl-2015:ACL:PPDB2.0}, we include the oracle result (the max Spearman's $\rho$ on the dataset), since this dataset was used for model selection for all other tasks, as well as test results where models were tuned on \annoppdb.
\vspace{-0.3cm}}
\end{table}

\subsection{\paragramphrase XXL} \label{sec:XXL}

\begin{table}[!h]
\setlength{\tabcolsep}{4pt}
\small
\centering
\begin{tabular} { |l || c | c | c || C{1.6cm} |} \hline
Dataset & 50\% & 75\% & Max & \paragramphrase-XXL\\
\hline
MSRpar & 51.5 & 57.6 & 73.4 & 44.8 \\
MSRvid & 75.5 & 80.3 & 88.0 & 79.6 \\
SMT-eur & 44.4 & 48.1 & 56.7 & 49.5 \\
OnWN & 60.8 & 65.9 & 72.7  & 70.4 \\
SMT-news & 40.1 & 45.4 & 60.9  & \textbf{63.3} \\
\hline
STS 2012 Average & 54.5 & 59.5 & 70.3 & 61.5 \\
\hline
headline & 64.0 & 68.3 & 78.4 & 73.9 \\
OnWN & 52.8 & 64.8 & 84.3 & 73.8 \\
FNWN & 32.7 & 38.1 & 58.2 & 47.7 \\
SMT & 31.8 & 34.6 & 40.4 & \textbf{40.4} \\
\hline
STS 2013 Average & 45.3 & 51.4 & 65.3 & 58.9 \\
\hline
deft forum & 36.6 & 46.8 & 53.1 & \textbf{53.4} \\
deft news & 66.2 & 74.0 & 78.5 & 74.4 \\
headline & 67.1 & 75.4 & 78.4 & 71.5 \\
images & 75.6 & 79.0 & 83.4 & 80.4 \\
OnWN &  78.0 & 81.1 & 87.5 & 81.5 \\
tweet news & 64.7 & 72.2 & 79.2 & 77.4 \\
\hline
STS 2014 Average & 64.7 & 71.4 & 76.7 & 73.1 \\
\hline
answers-forums & 61.3 & 68.2 & 73.9 & 69.1 \\
answers-students & 67.6 & 73.6 & 78.8 & 78.0 \\
belief & 67.7 & 72.2 & 77.2  & \textbf{78.2} \\
headline & 74.2 & 80.8 & 84.2 & 76.4 \\
images &  80.4 & 84.3 & 87.1 & 83.4 \\
\hline
STS 2015 Average & 70.2 & 75.8 & 80.2 & 77.0 \\
\hline
2014 SICK$^\ast$ &  71.4 & 79.9 & 82.8 & 72.7 \\
\hline
2015 Twitter & 49.9 & 52.5 & 61.9 & 52.4 \\
\hline
\end{tabular}
\caption{\label{fig:stats}
Results on SemEval textual similarity datasets (Pearson's $r  \times 100$) for \paragramphrase XXL embeddings. Results that match or exceed the best shared task system are shown in bold. $^\ast$For the 2014 SICK task, the median, $75^\textrm{th}$ percentile, and maximum include only the primary runs as the full set of results was not available.
}
\end{table}

Since we found that \paragramphrase embeddings have such strong performance, we trained this model on more data from PPDB and also used more data for hyperparameter tuning. For tuning, we used all of PPDB XL and trained for 10 epochs, then trained our final model for 10 epochs on the entire phrase section of PPDB XXL, consisting of 9,123,575 unique phrase pairs.\footnote{We fixed batchsize to 100 and $\delta$ to 0.4, as these were the optimal values for the experiment in Table~\ref{fig:STS results}. Then, for $\lambda_w$ we searched over $\{10^{-6}, 10^{-7}, 10^{-8}\}$, and tuned over MIX and MAX sampling. To optimize, we used AdaGrad with a learning rate of 0.05.} 
We show the results of this improved model, which we call \paragramphrase XXL, in Table~\ref{fig:stats}. We also report the median, $75^\textrm{th}$ percentile, and maximum score from our suite of textual similarity tasks. \paragramphrase XXL matches or exceeds the best performance on 4 of the datasets (SMT-news, SMT, deft forum, and belief) and is within 3 points of the best performance on 8 out of 22. 
We have made this trained model 
available to the research community.\footnote{Available at \url{http://ttic.uchicago.edu/~wieting}.} 

\subsection{Using Representations in Learned Models} \label{sec:supervised}

We explore two natural questions regarding our representations learned from PPDB: (1) can these embeddings improve the performance of other models through initialization and regularization? (2) can they effectively be used as features for downstream tasks? To address these questions, we used three tasks: The SICK similarity task, the SICK entailment task, and the Stanford Sentiment Treebank (SST) binary classification task~\citep{socher-13}. For the SICK similarity task, we minimize the objective function\footnote{This objective function has been shown to perform very strongly on text similarity tasks, significantly better than squared or absolute error.} from \cite{tai2015improved}. Given a score for a sentence pair in the range $[1, K]$, where $K$ is an integer, with sentence representations $h_L$ and $h_R$, and model parameters $\theta$, they first compute:
\begin{align}
h_\times &= h_L \odot h_R, \label{eq:sim-network} \:\:h_+ = |h_L - h_R|, \nonumber \\
h_s &= \sigma\left(W^{(\times)} h_\times  + W^{(+)} h_+ + b^{(h)} \right), \nonumber \\
\hat{p}_\theta     &= \mathrm{softmax}\left(W^{(p)} h_s + b^{(p)} \right), \nonumber \\
\hat{y}     &= r^T \hat{p}_\theta, \nonumber
\end{align}
where $r^T = [1~2~\dots~K]$. They then define a sparse target distribution $p$ that satisfies $y = r^T p$:
\begin{equation}
p_i = \begin{cases} y - \lfloor y \rfloor, & i = \lfloor y \rfloor + 1 \\ \lfloor y \rfloor - y + 1, & i = \lfloor y \rfloor  \\ 0 & \text{otherwise}\end{cases}
\end{equation}
for $1 \leq i \leq K$. Then they use the following loss, the regularized KL-divergence between $p$ and $\hat{p}_\theta$:
\begin{equation}
 J(\theta) = \frac{1}{m} \sum_{k=1}^m \mathrm{KL}\Big(p^{(k)}~\Big\|~\hat{p}^{(k)}_\theta\Big),\label{eq:KL}
\end{equation}
where $m$ is the number of training pairs and where we always use L$_2$ regularization on all compositional parameters\footnote{Word embeddings are regularized toward their initial state.} but omit these terms for clarity.  

We use nearly the same model for the entailment task, with the only differences being that the final softmax layer has three outputs and the cost function is the negative log-likelihood of the class labels. For sentiment, since it is a binary sentence classification task, we first encoded the sentence and then used a fully-connected layer with a sigmoid activation
followed by a softmax layer with two outputs. We used negative log-likelihood of the class labels as the cost function. All models use L$_2$ regularization on all parameters, except for the word embeddings, which are regularized back to their initial values with an L$_2$ penalty. 

We first investigated how these models performed in the standard setting, without using any models trained using PPDB data. We tuned hyperparameters on the development set of each dataset\footnote{For all models, we tuned batch-size over $\{25,50,100\}$, output dimension over $\{50,150,300\}$, $\lambda_c$ over $\{10^{-3},10^{-4},10^{-5},10^{-6}\}$, $\lambda_s = \lambda_c$, and $\lambda_w$ over  $\{10^{-3},10^{-4},10^{-5},10^{-6},10^{-7},10^{-8}\}$ as well as the option of not updating the embeddings for all models except the word averaging model. We again fix the output dimensionalities of all models which require this specification, to the dimensionality of our word embeddings (300). Additionally, for the classic RNN, we further tuned whether to use $\tanh$ or rectified linear unit activation functions; for the DANs we tuned over activation function ($\tanh$ or rectified linear unit) and the number of layers (either 1 or 2).} as well as on two optimization schemes: AdaGrad with learning rate of 0.05 and Adam with a learning rate of 0.001. We trained the models for 10 epochs and initialized the word embeddings with \paragramsl embeddings.

\begin{table}
\footnotesize
\centering
  \begin{tabular}{| l || c | c | c | c | c | c || c |}
    \hline
    \multirow{2}{*}{Task} & word & \multirow{2}{*}{proj.} & \multirow{2}{*}{DAN} & \multirow{2}{*}{RNN} & LSTM & LSTM & w/ {\it universal} \\
      & averaging &  &  &  & (no o.g.) & (o.g.) & regularization \\
    \hline
    similarity (SICK) & \bf 86.40 &  85.93 & 85.96 & 73.13 & 85.45 & 83.41 &  \bf 86.84\\
    entailment (SICK) & \bf 84.6 & 84.0 & 84.5 & 76.4 & 83.2 & 82.0 & \bf 85.3\\
    binary sentiment (SST) & 83.0 & 83.0 & 83.4 & 86.5  & 86.6 & \bf 89.2 & 86.9\\
    \hline
  \end{tabular}
  \caption{\label{fig:Supervised}
  Results from supervised training of each compositional architecture on similarity, entailment, and sentiment tasks. The last column shows results regularizing to our {\it universal} parameters from the models in Table~\ref{fig:STS results}. The first row shows Pearson's $r \times 100$ and the last two show accuracy. 
\vspace{-0.3cm}}
\end{table}

The results are shown in Table~\ref{fig:Supervised}. We find that using word averaging as the compositional architecture outperforms the other architectures for similarity and entailment. 
However, for sentiment classification, the LSTM is much stronger than the averaging models. This suggests that the superiority of a compositional architecture can vary widely depending on the evaluation, and motivates future work to compare these architectures on additional tasks. 

These results are very competitive with the state of the art 
on these tasks. Recent strong results on the SICK similarity task include 86.86 using a convolutional neural network~\citep{he-gimpel-lin:2015:EMNLP} and 86.76 using a tree-LSTM~\citep{tai2015improved}. 
For entailment, the best result we are aware of is 85.1~\citep{beltagy2015representing}. On sentiment, the best previous result is 88.1~\citep{kim-14}, which our LSTM surprisingly outperforms by a significant margin. 
\rev{We note that these experiments simply compare compositional architectures using only the provided training data for each task, tuning on the respective development sets. We did not use any PPDB data for these results, other than that used to train the initial \paragramsl embeddings. Our results appear to show that standard neural architectures can perform surprisingly well given strong word embeddings and thorough tuning over the hyperparameter space.}

\subsubsection{Regularization and Initialization to Improve Textual Similarity Models} \label{sec:reg}

In this setting, we initialize each respective model to the parameters learned from PPDB (calling them {\it universal} parameters) 
and augment Eq.~\refeq{eq:KL} 
with three separate regularization terms with the following weights: $\lambda_s$ which regularizes the classification parameters (the two layers used in the classification step after obtaining representations), $\lambda_w$ for regularizing the word parameters toward the learned $W_w$ from PPDB, and $\lambda_{c}$ for regularizing the compositional parameters (for all models except for the word averaging model) back to their initial values.\footnote{We tuned $\lambda_s$ over $\{10^{-3},10^{-4},10^{-5},10^{-6}\}$, $\lambda_c$ over $\{10^{-2},10^{-3},10^{-4},10^{-5},10^{-6}\}$, and $\lambda_w$ over $\{10^{-3},10^{-4},10^{-5},10^{-6},10^{-7},10^{-8}\}$. All other hyperparameters were tuned as previously described.}  In all cases, we regularize to the universal parameters using L$_2$ regularization.

The results are shown in the last column of Table~\ref{fig:Supervised}, and we only show results for the best performing models on each task (word averaging for similarity/entailment, LSTM with output gate for sentiment). 
Interestingly, it seems that regularizing to our universal parameters significantly improves results for the similarity and entailment tasks which are competitive or better than the state-of-the-art, but harms the LSTM's performance on the sentiment classification task.

\subsubsection{Representations as Features} \label{sec:rep}

\begin{table}[h!]
\small
\centering
  \begin{tabular}{| l || c | c | c | c | c |}
    \hline
    \multirow{2}{*}{Task}& \multicolumn{3}{c|}{\paragramphrase} & \multicolumn{2}{c|}{skip-thought} \\
 &  300 & 1200 & 2400 & uni-skip & bi-skip\\
    \hline
    similarity (SICK) & 82.15 & 82.85 & \bf 84.94 & 84.77 & 84.05 \\
    entailment (SICK)  & 80.2 & 80.1 & \bf 83.1 & - & -\\
    binary sentiment (SST)  & \bf 79.7 & 78.8 & 79.4 & - & -\\
    \hline
  \end{tabular}
  \caption{\label{fig:Supervised-Rep}
Results from supervised training on similarity, entailment, and sentiment tasks, except that we keep the sentence representations fixed to our \paragramphrase model.
The first row shows Pearson's $r \times 100$ and the last two show accuracy, with boldface showing the highest score in each row. 
\vspace{-0.3cm}}
\end{table}

We also investigate how our \paragramphrase embeddings perform as features for supervised tasks. We use a similar set-up as in \cite{kiros2015skip} and encode the sentences by averaging our \paragramphrase embeddings and then just learn the classification parameters without updating the embeddings. To provide a more apt comparison to skip-thought vectors, we also learned a linear projection matrix to increase dimensionality of our \paragramphrase embeddings. 
We chose 1200 and 2400 dimensions in order to both see the dependence of dimension on performance, and so that they can be compared fairly with skip-thought vectors. Note that 2400 dimensions is the same dimensionality as the uni-skip and bi-skip models in \cite{kiros2015skip}.

The 300 dimension case corresponds to the \paragramphrase embeddings from Table~\ref{fig:STS results}. We tuned our higher dimensional models on PPDB as described previously in Section~\ref{sec:results} before training on PPDB XL.\footnote{Note that we fixed batch-size to 100, $\delta$ to 0.4, and used MAX sampling as these were the optimal parameters for the \paragramphrase embeddings. We tuned the other hyperparameters as described in Section~\ref{sec:results} with the exception of $\lambda_c$ which was tuned over $\{10^{-4},10^{-5},10^{-6}, 10^{-7}, 10^{-8}\}$.} Then we trained the same models for the similarity, entailment, and sentiment tasks as described in Section~\ref{sec:supervised} for 20 epochs. We again tuned $\lambda_s$ over $\{10^{-3},10^{-4},10^{-5},10^{-6}\}$ and tuned over the two optimization schemes of AdaGrad with learning rate of 0.05 and Adam with a learning rate of 0.001. Note that we are not updating the word embeddings or the projection matrix during training. 

The results are shown in Table~\ref{fig:Supervised-Rep}.
The similarity and entailment tasks show clear improvements as we project the embeddings into the 2400 dimensional space. In fact, our results outperform both types of skip-thought embeddings on the single task that we overlap. 
However, the sentiment task does not benefit from higher dimensional representations, 
which is consistent with our regularization experiments in which sentiment also did not show improvement. Therefore, it seems that our models learned from PPDB are more effective for {\it similarity} tasks than {\it classification} tasks, but this hypothesis requires further investigation.

\section{Discussion} \label{sec:discussion}
It is interesting that the LSTM, with or without output gates, is outperformed by much simpler models on the similarity and entailment tasks studied in this paper. 
We now consider possible explanations for this trend.

The first hypothesis we test is based on length. Since PPDB contains short text snippets of a few words, the LSTM may not know how to handle the longer sentences that occur in our evaluation tasks. If this is true, the LSTM 
would perform much better on short text snippets and its performance would degrade as their length increases. To test this hypothesis, we took all 12,108 pairs 
from the 20 SemEval STS tasks and binned them by length.\footnote{For each pair, we computed the number of tokens in each of the two pieces of text, took the max, and then binned based on this value.}  We then computed the Pearson's $r$ for each bin. 
The results are shown in Table~\ref{fig:length} and show that while the LSTM models do perform better on the shortest text pairs, they are still outperformed, at all lengths, by the \paragramphrase 
model.\footnote{Note that for the analysis in Sections~\ref{sec:discussion} and~\ref{sec:qual}, the models used were selected from earlier experiments. They are not the same as those used to obtain the results in Table~\ref{fig:STS results}.}

\begin{table}[h!]
\small
\centering
  \begin{tabular}{| C{1cm} | C{1.5cm} | C{1.2cm} | C{1.2cm} | C{1.5cm} |}
    \hline
    Max Length & \paragramphrase  & LSTM (no o.g.) & LSTM (o.g.) & \paragramsl\\
    \hline
$ \leq 4$ & \bf 72.7 & 63.4 & 58.8 & 66.3\\
5 & \bf 74.0  & 54.5 & 48.4 & 65.0\\
6 & \bf 70.5 & 52.6 & 48.2 & 50.1\\
7 & \bf 73.7 & 56.9 & 50.6 & 56.4\\
8 & \bf 75.5 & 60.2 & 52.4 & 60.1\\
9 & \bf 73.0  & 58.0 & 48.8 & 58.8\\
$ \geq 10$ & \bf 72.6 & 55.6 & 53.8 & 58.4\\
    \hline
  \end{tabular}
  \caption{\label{fig:length}
Performance (Pearson's $r  \times 100$) as a function of the maximum number of tokens in the sentence pairs over all 20 SemEval STS datasets.
}
\end{table}

We next consider whether the LSTM has worse generalization due to overfitting on the training data. To test this, we analyzed how the models performed on the training data (PPDB XL) by computing the average difference between the cosine similarity of the gold phrase pairs and the negative examples.\footnote{More precisely, for each gold pair $\langle g_1, g_2\rangle$, and $n_i$, the respective negative example of each $g_i$, we computed $2\cdot \cos(g_1,g_2) - \cos(n_1,g_1) - \cos(n_2,g_2)$ and averaged this value over all pairs.} We found that all models had very similar scores: 0.7535, 0.7572, 0.7565, and 0.7463 for \paragramphrase, projection, LSTM (o.g.), and LSTM (no o.g.). This, along with the similar performance of the models on the PPDB tasks in Table~\ref{fig:PPDB results}, suggests that overfitting is not the cause of the worse performance of the LSTM model.

Lastly, we consider 
whether the LSTM's weak performance was a result of insufficient tuning or optimization. We first note that we actually ran more hyperparameter tuning experiments for the LSTM models than either the \paragramphrase or projection models, since we tuned the decision to use an output gate. Secondly, we note that \cite{tai2015improved} had a similar LSTM result on the SICK dataset (Pearson's $r$ of 85.28 to our 85.45) to show that our LSTM implementation/tuning procedure is able to match or exceed performance of another published LSTM result. 
Thirdly, the similar performance across models on the PPDB tasks (Table~\ref{fig:PPDB results}) suggests that no model had a large advantage during tuning; all found hyperparameters that comfortably beat the \paragramsl addition baseline. Finally, we point out that we tuned over learning rate and optimization strategy, as well as experimented with clipping gradients, in order to rule out optimization issues. 

\subsection{Under-Trained Embeddings}

One limitation of our new \paragramphrase vectors is that many of our embeddings are under-trained. The number of unique tokens occurring in our training data, PPDB XL, is 37,366. However, the number of tokens appearing more than 100 times is just 7,113. Thus, one clear source of improvement for our model would be to address under-trained embeddings for tokens appearing in our test data.

In order to gauge the effect under-trained embeddings and unknown words have on our model, we calculated the fraction of words in each of our 22 SemEval datasets that do not occur at least 100 times in PPDB XL along with our performance deviation from the $75^\textrm{th}$ percentile of each dataset. We found that this fraction had a Spearman's $\rho$ of -45.1 with the deviation from the $75^\textrm{th}$ percentile indicating that there is a significant negative correlation between the fraction of OOV words and performance on these STS tasks.

\subsection{Using More PPDB} \label{performance_vs_training}

\subsubsection{Performance Versus Amount of Training Data}
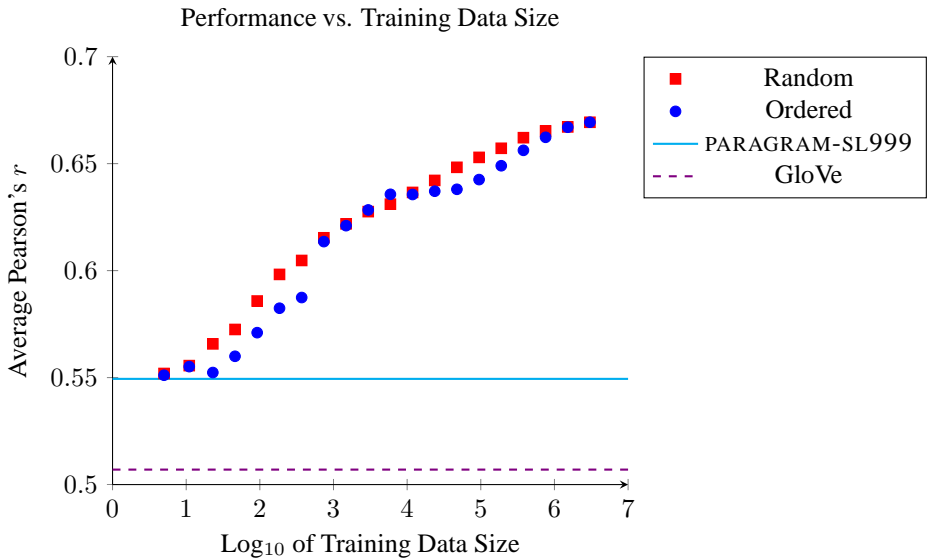
\begin{figure}[h!]
\centering
\begin{tikzpicture}
\begin{axis}[
    axis lines = left,
    xlabel = $x$,
    ylabel = {$f(x)$},
    ymin = 0.5,
    ymax = 0.7,
    xlabel = Log$_{10}$ of Training Data Size,
    ylabel = Average Pearson's $r$,
    legend pos= outer north east,
    title = Performance vs. Training Data Size,
]
\addplot [
    color=red,
    mark = square*,
    only marks
]
table[x=x,y=y,col sep=comma] from {random_points.txt};
\addlegendentry{Random}
\addplot [
    color=blue,
    mark = *,
    only marks
]
table[x=x,y=y,col sep=comma] from {ordered_points.txt};
\addlegendentry{Ordered}
\addplot [
    domain=0:7, 
    samples=100, 
    color=cyan,
    thick
    ]
    {0.5494};
\addlegendentry{\paragramsl}
\addplot [
    domain=0:7, 
    samples=100, 
    color=violet,
    thick,
    dashed,
    ]
    {0.5070};
\addlegendentry{GloVe} 
\end{axis}
\end{tikzpicture}
 \caption{Performance of the \paragramphrase embeddings as measured by the average Pearson's $r$ on 22 textual similarity datasets versus the amount of training data from PPDB on a log scale. Each datapoint contains twice as much training data as the previous one. Random and Ordered refer to whether we shuffled the XL paraphrase pairs from PPDB or kept them in order. We also show baselines of averaging \paragramsl and GloVe embeddings. 
}
\label{fig:perf}
\end{figure}

Models in related work such as \cite{kiros2015skip} and \cite{li2015hierarchical} require significant training time on GPUs, on the order of multiple weeks. Moreover, 
dependence of model performance upon training data size 
is unclear. 
To investigate this dependence for our \paragramphrase model, we trained on different amounts of data and plotted the performance. The results are shown in Figure~\ref{fig:perf}. 
We start with PPDB XL which has 3,033,753 
unique phrase pairs
and then divide by two 
until there are fewer than 10 phrase pairs.\footnote{The smallest dataset contained 5 pairs.} For each data point (each division by two), we trained a model with that number of phrase pairs for 10 epochs. We use the average Pearson correlation for all 22 datasets in Table~\ref{fig:STS results} as the dependent variable in our plot. 

We experimented with two different ways of selecting training data. 
The first (``Ordered'')
retains the order of the phrase pairs in PPDB, which ensures the smaller datasets contain higher confidence phrase pairs. 
The second (``Random'') randomly permutes 
PPDB XL before constructing the smaller datasets. In both methods, each larger dataset contains the previous one plus as many new phase pairs.

We make three observations about the plot in Figure~\ref{fig:perf}. 
The first is that performance continually increases as more training data is added. This is encouraging as our embeddings can continually improve with more data. Secondly, we note the sizable 
improvement (4 points) over the \paragramsl baseline by training on just 92 phrase pairs from PPDB. 
Finally, we note the difference between 
randomly permuting the training data and using the order from PPDB (which reflects the confidence that the phrases in each pair possess the paraphrase relationship). 
Performance of the randomly permuted data is usually slightly better than that of the ordered data, until the performance gap vanishes once half of PPDB XL is used. We suspect this behavior is due to the {\it safe} phrase pairs that occur in the beginning of PPDB. These high-confidence phrase pairs usually have only slight differences and therefore are not as useful for training our model.

\section{Qualitative Analysis} \label{sec:qual}
\begin{table}[h!]
	\centering
	\scriptsize
	\begin{tabular}{|l|l|l|}
		\hline
		Word & \paragramphrase Nearest Neighbors & \paragramsl Nearest Neighbors\\
		\hline
		unlike & contrary, contrast, opposite, versa, conversely, opposed, contradiction &  than, although, whilst, though, albeit, kinda, alike\\
		2 &  2.0, two, both, ii, 2nd, couple, 02  & 2.0, 3, 1, b, ii, two, 2nd\\
		ladies & girls, daughters, honorable, females, girl, female, dear  &  gentlemen, colleague, fellow, girls, mr, madam, dear\\
		lookin & staring, looking, watching, look, searching, iooking, seeking &  doin, goin, talkin, sayin, comin, outta, somethin\\
		disagree & agree, concur, agreeing, differ, accept  &  disagreement, differ, dispute, difference, disagreements \\
		\hline
	\end{tabular}
	\caption{\label{fig:nn} Nearest neighbors of \paragramphrase and \paragramsl word embeddings sorted by cosine similarity.}
\end{table}
To explore other differences between our \paragramphrase vectors and the \paragramsl vectors that were used for initialization, we inspected lists of nearest neighbors in each vector space. 
When obtaining nearest neighbors, we restricted our search to the 10,000 most common tokens in PPDB XL 
to ensure that the \paragramphrase vectors were not too under-trained. Some informative neighbors are shown in Table~\ref{fig:nn}. In the first four rows, we see that the \paragramphrase embeddings have neighbors with a strong paraphrasing relationship. They tend to avoid having neighbors that are antonyms or co-hyponyms such as {\it unlike} and {\it alike} or {\it 2} and {\it 3} which are an issue for the \paragramsl embeddings. In contrast to the first four rows, the last row shows a problematic effect of our bag-of-words composition function: 
{\it agree} is the nearest neighbor of {\it disagree}. The reason for this is that there are numerous pairs in PPDB XL such as {\it i disagree} and {\it i do not agree} that encourage {\it disagree} and {\it agree} to have high cosine similarity. A model that takes context into account could resolve this issue. The difficulty would be finding a model that does so while still generalizing well, as we found that our \paragramphrase embeddings generalize better than learning a weight matrix or using a recurrent neural network. We leave this for future work. 

When we take a closer look at our \paragramphrase embeddings, we find that information-bearing content words, such as {\it poverty}, {\it kidding}, {\it humanitarian}, {\it 18}, and {\it july} have 
the largest L$_2$ norms, while 
words such as {\it of}, {\it it}, {\it to}, {\it hereby} and {\it the} have the smallest. 
\rev{\citet{pham-EtAl:2015:ACL-IJCNLP} noted this same phenomenon in their closely-related compositional model. 
Interestingly, we found that this weighting} explains much of the success of our model. In order to quantify exactly how much, we calculated a weight for each token in our working vocabulary\footnote{This corresponds to the 42,091 tokens that appear in the intersection of our \paragramsl vocabulary, the test sets of all STS tasks in our evaluation, and PPDB XL plus an unknown word token.} simply by summing up the absolute value of all components of its \paragramphrase vector. Then we multiplied each weight by its corresponding \paragramsl word vector.  We computed the average Pearson's $r$ over all 22 datasets in Table~\ref{fig:STS results}. The \paragramsl vectors have an average correlation of 54.94, the \paragramphrase vectors have 66.83, and the scaled \paragramsl vectors, where each is multiplied by its computed weight, have an average Pearson's $r$ of 62.64. Therefore, it can be surmised that at least 64.76\% of the improvement over the initial \paragramsl vectors is due to weighting tokens by their importance.\footnote{We also trained a model in which we only a learn a single multiplicative parameter for each word in our vocabulary, keeping the word embeddings fixed to the \paragramsl embeddings. We trained for 10 epochs on all phrase pairs in PPDB XL. The resulting average Pearson's $r$, after tuning on the Pavlick et al.~PPDB task, was 62.06, which is slightly lower than using the absolute value of each \paragramphrase vector as its multiplicative weight.}  

We also investigated the connection between these multiplicative weights and word frequency. 
To do so, we calculated the frequency of all tokens in PPDB XL.\footnote{Tokens that did not appear in PPDB XL were assigned a frequency of 1.} We then normalized these by the total number of tokens in PPDB XL and used the reciprocal of these scores as the multiplicative weights. Thus less frequent words have more weight than more frequent words. 
With this baseline weighting method, the average Pearson's $r$ is 45.52, indicating that the weights we obtain for these words are more sophisticated than mere word frequency. These weights are 
potentially useful for other applications that can benefit from modeling 
word importance, 
such as information retrieval.

\section{Conclusion}

We introduced an approach to create universal sentence embeddings and propose our model as the new baseline for embedding sentences, as it is simple, efficient, and performs strongly across a broad range of tasks and domains. 
Moreover, our representations do not require the use of any neural network architecture. The embeddings can be simply averaged for a given sentence in an NLP application to create its sentence embedding. We also find that our representations can improve general text similarity and entailment models when used as a prior and can achieve strong performance even when used as fixed representations in a classifier. Future work will focus on improving our embeddings by effectively handling undertrained words as well as by exploring new models that generalize even better to the large suite of text similarity tasks used in our experiments.

\subsubsection*{Acknowledgments}

We would like to thank Yoon Kim, the anonymous reviewers, and the area chair for their valuable comments. We would also like to thank the  developers of Theano~\citep{bergstra+al:2010-scipy, Bastien-Theano-2012} and thank NVIDIA Corporation for donating GPUs used in this research.

\bibliography{iclr2016_conference}

\begin{thebibliography}{77}
\providecommand{\natexlab}[1]{#1}
\providecommand{\url}[1]{\texttt{#1}}
\expandafter\ifx\csname urlstyle\endcsname\relax
  \providecommand{\doi}[1]{doi: #1}\else
  \providecommand{\doi}{doi: \begingroup \urlstyle{rm}\Url}\fi

\bibitem[Agirre et~al.(2012)Agirre, Diab, Cer, and
  Gonzalez-Agirre]{agirre2012semeval}
Agirre, Eneko, Diab, Mona, Cer, Daniel, and Gonzalez-Agirre, Aitor.
\newblock {SemEval}-2012 task 6: A pilot on semantic textual similarity.
\newblock In \emph{Proceedings of the First Joint Conference on Lexical and
  Computational Semantics-Volume 1: Proceedings of the main conference and the
  shared task, and Volume 2: Proceedings of the Sixth International Workshop on
  Semantic Evaluation}. Association for Computational Linguistics, 2012.

\bibitem[Agirre et~al.(2013)Agirre, Cer, Diab, Gonzalez-Agirre, and
  Guo]{diab2013eneko}
Agirre, Eneko, Cer, Daniel, Diab, Mona, Gonzalez-Agirre, Aitor, and Guo,
  Weiwei.
\newblock *{SEM} 2013 shared task: Semantic textual similarity.
\newblock In \emph{Second Joint Conference on Lexical and Computational
  Semantics (*{SEM}), Volume 1: Proceedings of the Main Conference and the
  Shared Task: Semantic Textual Similarity}, 2013.

\bibitem[Agirre et~al.(2014)Agirre, Banea, Cardie, Cer, Diab, Gonzalez-Agirre,
  Guo, Mihalcea, Rigau, and Wiebe]{agirre2014semeval}
Agirre, Eneko, Banea, Carmen, Cardie, Claire, Cer, Daniel, Diab, Mona,
  Gonzalez-Agirre, Aitor, Guo, Weiwei, Mihalcea, Rada, Rigau, German, and
  Wiebe, Janyce.
\newblock {SemEval}-2014 task 10: Multilingual semantic textual similarity.
\newblock In \emph{Proceedings of the 8th International Workshop on Semantic
  Evaluation ({SemEval} 2014)}, 2014.

\bibitem[Agirre et~al.(2015)Agirre, Banea, Cardie, Cer, Diab, Gonzalez-Agirre,
  Guo, Lopez-Gazpio, Maritxalar, Mihalcea, Rigau, Uria, and
  Wiebe]{agirre2015semeval}
Agirre, Eneko, Banea, Carmen, Cardie, Claire, Cer, Daniel, Diab, Mona,
  Gonzalez-Agirre, Aitor, Guo, Weiwei, Lopez-Gazpio, Inigo, Maritxalar, Montse,
  Mihalcea, Rada, Rigau, German, Uria, Larraitz, and Wiebe, Janyce.
\newblock {SemEval}-2015 task 2: Semantic textual similarity, {English},
  {Spanish} and pilot on interpretability.
\newblock In \emph{Proceedings of the 9th International Workshop on Semantic
  Evaluation ({SemEval} 2015)}, 2015.

\bibitem[Bansal et~al.(2014)Bansal, Gimpel, and Livescu]{bansal2014tailoring}
Bansal, Mohit, Gimpel, Kevin, and Livescu, Karen.
\newblock Tailoring continuous word representations for dependency parsing.
\newblock In \emph{Proceedings of the Annual Meeting of the Association for
  Computational Linguistics}, 2014.

\bibitem[Baroni et~al.(2014)Baroni, Bernardi, and Zamparelli]{baroni2014frege}
Baroni, Marco, Bernardi, Raffaela, and Zamparelli, Roberto.
\newblock Frege in space: A program of compositional distributional semantics.
\newblock \emph{Linguistic Issues in Language Technology}, 9, 2014.

\bibitem[Bastien et~al.(2012)Bastien, Lamblin, Pascanu, Bergstra, Goodfellow,
  Bergeron, Bouchard, and Bengio]{Bastien-Theano-2012}
Bastien, Fr{\'{e}}d{\'{e}}ric, Lamblin, Pascal, Pascanu, Razvan, Bergstra,
  James, Goodfellow, Ian~J., Bergeron, Arnaud, Bouchard, Nicolas, and Bengio,
  Yoshua.
\newblock Theano: new features and speed improvements, 2012.

\bibitem[Belinkov \& Glass(2015)Belinkov and Glass]{belinkov-glass:2015:EMNLP}
Belinkov, Yonatan and Glass, James.
\newblock Arabic diacritization with recurrent neural networks.
\newblock In \emph{Proceedings of the 2015 Conference on Empirical Methods in
  Natural Language Processing}, 2015.

\bibitem[Beltagy et~al.(2015)Beltagy, Roller, Cheng, Erk, and
  Mooney]{beltagy2015representing}
Beltagy, Islam, Roller, Stephen, Cheng, Pengxiang, Erk, Katrin, and Mooney,
  Raymond~J.
\newblock Representing meaning with a combination of logical form and vectors.
\newblock \emph{arXiv preprint arXiv:1505.06816}, 2015.

\bibitem[Bergstra et~al.(2010)Bergstra, Breuleux, Bastien, Lamblin, Pascanu,
  Desjardins, Turian, Warde-Farley, and Bengio]{bergstra+al:2010-scipy}
Bergstra, James, Breuleux, Olivier, Bastien, Fr{\'{e}}d{\'{e}}ric, Lamblin,
  Pascal, Pascanu, Razvan, Desjardins, Guillaume, Turian, Joseph, Warde-Farley,
  David, and Bengio, Yoshua.
\newblock Theano: a {CPU} and {GPU} math expression compiler.
\newblock In \emph{Proceedings of the Python for Scientific Computing
  Conference ({SciPy})}, June 2010.

\bibitem[Bird et~al.(2009)Bird, Klein, and Loper]{bird2009natural}
Bird, Steven, Klein, Ewan, and Loper, Edward.
\newblock \emph{Natural language processing with Python}.
\newblock O'Reilly Media, Inc., 2009.

\bibitem[Blacoe \& Lapata(2012)Blacoe and Lapata]{Blacoe2012}
Blacoe, William and Lapata, Mirella.
\newblock A comparison of vector-based representations for semantic
  composition.
\newblock In \emph{Proceedings of the 2012 Joint Conference on Empirical
  Methods in Natural Language Processing and Computational Natural Language
  Learning}, 2012.

\bibitem[Bordes et~al.(2014{\natexlab{a}})Bordes, Chopra, and Weston]{D14-1067}
Bordes, Antoine, Chopra, Sumit, and Weston, Jason.
\newblock Question answering with subgraph embeddings.
\newblock In \emph{Proceedings of the 2014 Conference on Empirical Methods in
  Natural Language Processing (EMNLP)}, 2014{\natexlab{a}}.

\bibitem[Bordes et~al.(2014{\natexlab{b}})Bordes, Weston, and
  Usunier]{bordes2014open}
Bordes, Antoine, Weston, Jason, and Usunier, Nicolas.
\newblock Open question answering with weakly supervised embedding models.
\newblock In \emph{Machine Learning and Knowledge Discovery in Databases}.
  Springer, 2014{\natexlab{b}}.

\bibitem[Chen et~al.(2015{\natexlab{a}})Chen, Qiu, Zhu, Liu, and
  Huang]{chen-EtAl:2015:EMNLP2}
Chen, Xinchi, Qiu, Xipeng, Zhu, Chenxi, Liu, Pengfei, and Huang, Xuanjing.
\newblock Long short-term memory neural networks for {Chinese} word
  segmentation.
\newblock In \emph{Proceedings of the 2015 Conference on Empirical Methods in
  Natural Language Processing}, 2015{\natexlab{a}}.

\bibitem[Chen et~al.(2015{\natexlab{b}})Chen, Qiu, Zhu, Wu, and
  Huang]{chen-EtAl:2015:EMNLP1}
Chen, Xinchi, Qiu, Xipeng, Zhu, Chenxi, Wu, Shiyu, and Huang, Xuanjing.
\newblock Sentence modeling with gated recursive neural network.
\newblock In \emph{Proceedings of the 2015 Conference on Empirical Methods in
  Natural Language Processing}, 2015{\natexlab{b}}.

\bibitem[Collobert et~al.(2011)Collobert, Weston, Bottou, Karlen, Kavukcuoglu,
  and Kuksa]{Collobert:2011:NLP}
Collobert, Ronan, Weston, Jason, Bottou, L{\'e}on, Karlen, Michael,
  Kavukcuoglu, Koray, and Kuksa, Pavel.
\newblock Natural language processing (almost) from scratch.
\newblock \emph{J. Mach. Learn. Res.}, 12, 2011.

\bibitem[Duchi et~al.(2011)Duchi, Hazan, and Singer]{Duchi}
Duchi, John, Hazan, Elad, and Singer, Yoram.
\newblock Adaptive subgradient methods for online learning and stochastic
  optimization.
\newblock \emph{J. Mach. Learn. Res.}, 12, 2011.

\bibitem[Dzikovska et~al.(2010)Dzikovska, Moore, Steinhauser, Campbell, Farrow,
  and Callaway]{dzikovska2010beetle}
Dzikovska, Myroslava~O, Moore, Johanna~D, Steinhauser, Natalie, Campbell,
  Gwendolyn, Farrow, Elaine, and Callaway, Charles~B.
\newblock Beetle {II}: a system for tutoring and computational linguistics
  experimentation.
\newblock In \emph{Proceedings of the ACL 2010 System Demonstrations}, 2010.

\bibitem[Faruqui \& Dyer(2014)Faruqui and Dyer]{faruqui-dyer:2014:EACL}
Faruqui, Manaal and Dyer, Chris.
\newblock Improving vector space word representations using multilingual
  correlation.
\newblock In \emph{Proceedings of the 14th Conference of the European Chapter
  of the Association for Computational Linguistics}, 2014.

\bibitem[Filippova et~al.(2015)Filippova, Alfonseca, Colmenares, Kaiser, and
  Vinyals]{filippova-EtAl:2015:EMNLP}
Filippova, Katja, Alfonseca, Enrique, Colmenares, Carlos~A., Kaiser, Lukasz,
  and Vinyals, Oriol.
\newblock Sentence compression by deletion with {LSTMs}.
\newblock In \emph{Proceedings of the 2015 Conference on Empirical Methods in
  Natural Language Processing}, 2015.

\bibitem[Finkelstein et~al.(2001)Finkelstein, Gabrilovich, Matias, Rivlin,
  Solan, Wolfman, and Ruppin]{finkelstein2001placing}
Finkelstein, Lev, Gabrilovich, Evgeniy, Matias, Yossi, Rivlin, Ehud, Solan,
  Zach, Wolfman, Gadi, and Ruppin, Eytan.
\newblock Placing search in context: The concept revisited.
\newblock In \emph{Proceedings of the 10th international conference on World
  Wide Web}. ACM, 2001.

\bibitem[Ganitkevitch et~al.(2013)Ganitkevitch, Durme, and
  Callison-Burch]{GanitkevitchDC13}
Ganitkevitch, Juri, Durme, Benjamin~Van, and Callison-Burch, Chris.
\newblock Ppdb: The paraphrase database.
\newblock In \emph{HLT-NAACL}. The Association for Computational Linguistics,
  2013.

\bibitem[Gers et~al.(2003)Gers, Schraudolph, and Schmidhuber]{gers2003learning}
Gers, Felix~A, Schraudolph, Nicol~N, and Schmidhuber, J{\"u}rgen.
\newblock Learning precise timing with lstm recurrent networks.
\newblock \emph{The Journal of Machine Learning Research}, 3, 2003.

\bibitem[Graves et~al.(2008)Graves, Liwicki, Bunke, Schmidhuber, and
  Fern\'{a}ndez]{NIPS2007_3213}
Graves, Alex, Liwicki, Marcus, Bunke, Horst, Schmidhuber, J\"{u}rgen, and
  Fern\'{a}ndez, Santiago.
\newblock Unconstrained on-line handwriting recognition with recurrent neural
  networks.
\newblock In \emph{Advances in Neural Information Processing Systems 20}. 2008.

\bibitem[Graves et~al.(2013)Graves, Mohamed, and Hinton]{graves2013speech}
Graves, Alex, Mohamed, Abdel-rahman, and Hinton, Geoffrey.
\newblock Speech recognition with deep recurrent neural networks.
\newblock In \emph{2013 IEEE International Conference on Acoustics, Speech and
  Signal Processing (ICASSP)}, 2013.

\bibitem[Greff et~al.(2015)Greff, Srivastava, Koutn{\'\i}k, Steunebrink, and
  Schmidhuber]{greff2015lstm}
Greff, Klaus, Srivastava, Rupesh~Kumar, Koutn{\'\i}k, Jan, Steunebrink, Bas~R,
  and Schmidhuber, J{\"u}rgen.
\newblock {LSTM}: A search space odyssey.
\newblock \emph{arXiv preprint arXiv:1503.04069}, 2015.

\bibitem[He et~al.(2015)He, Gimpel, and Lin]{he-gimpel-lin:2015:EMNLP}
He, Hua, Gimpel, Kevin, and Lin, Jimmy.
\newblock Multi-perspective sentence similarity modeling with convolutional
  neural networks.
\newblock In \emph{Proceedings of the 2015 Conference on Empirical Methods in
  Natural Language Processing}, 2015.

\bibitem[Hermann \& Blunsom(2014)Hermann and
  Blunsom]{hermann-blunsom:2014:P14-1}
Hermann, Karl~Moritz and Blunsom, Phil.
\newblock Multilingual models for compositional distributed semantics.
\newblock In \emph{Proceedings of the 52nd Annual Meeting of the Association
  for Computational Linguistics (Volume 1: Long Papers)}, 2014.

\bibitem[Hermann et~al.(2015)Hermann, Ko\v{c}isk\'y, Grefenstette, Espeholt,
  Kay, Suleyman, and Blunsom]{nips15_hermann}
Hermann, Karl~Moritz, Ko\v{c}isk\'y, Tom\'a\v{s}, Grefenstette, Edward,
  Espeholt, Lasse, Kay, Will, Suleyman, Mustafa, and Blunsom, Phil.
\newblock Teaching machines to read and comprehend.
\newblock In \emph{Advances in Neural Information Processing Systems}, 2015.

\bibitem[Hill et~al.(2015)Hill, Reichart, and Korhonen]{HillRK14}
Hill, Felix, Reichart, Roi, and Korhonen, Anna.
\newblock {SimLex}-999: Evaluating semantic models with (genuine) similarity
  estimation.
\newblock \emph{Computational Linguistics}, 41\penalty0 (4), 2015.

\bibitem[Hochreiter \& Schmidhuber(1997)Hochreiter and
  Schmidhuber]{hochreiter1997long}
Hochreiter, Sepp and Schmidhuber, J{\"u}rgen.
\newblock Long short-term memory.
\newblock \emph{Neural computation}, 9\penalty0 (8), 1997.

\bibitem[Hu et~al.(2014)Hu, Lu, Li, and Chen]{NIPS2014_5550}
Hu, Baotian, Lu, Zhengdong, Li, Hang, and Chen, Qingcai.
\newblock Convolutional neural network architectures for matching natural
  language sentences.
\newblock In \emph{Advances in Neural Information Processing Systems}, 2014.

\bibitem[Huang et~al.(2013)Huang, He, Gao, Deng, Acero, and
  Heck]{huang2013learning}
Huang, Po-Sen, He, Xiaodong, Gao, Jianfeng, Deng, Li, Acero, Alex, and Heck,
  Larry.
\newblock Learning deep structured semantic models for web search using
  clickthrough data.
\newblock In \emph{Proceedings of the 22nd ACM international conference on
  Conference on information \& knowledge management}, 2013.

\bibitem[\.Irsoy \& Cardie(2014)\.Irsoy and Cardie]{irsoy-drsv}
\.Irsoy, Ozan and Cardie, Claire.
\newblock Deep recursive neural networks for compositionality in language.
\newblock In \emph{Advances in Neural Information Processing Systems 27}. 2014.

\bibitem[Iyyer et~al.(2015)Iyyer, Manjunatha, Boyd-Graber, and
  Daum\'{e}~III]{iyyer-EtAl:2015:ACL-IJCNLP}
Iyyer, Mohit, Manjunatha, Varun, Boyd-Graber, Jordan, and Daum\'{e}~III, Hal.
\newblock Deep unordered composition rivals syntactic methods for text
  classification.
\newblock In \emph{Proceedings of the 53rd Annual Meeting of the Association
  for Computational Linguistics and the 7th International Joint Conference on
  Natural Language Processing (Volume 1: Long Papers)}, 2015.

\bibitem[Kalchbrenner et~al.(2014)Kalchbrenner, Grefenstette, and
  Blunsom]{kalchbrenner-grefenstette-blunsom:2014:P14-1}
Kalchbrenner, Nal, Grefenstette, Edward, and Blunsom, Phil.
\newblock A convolutional neural network for modelling sentences.
\newblock In \emph{Proceedings of the 52nd Annual Meeting of the Association
  for Computational Linguistics (Volume 1: Long Papers)}, 2014.

\bibitem[Kim(2014)]{kim-14}
Kim, Yoon.
\newblock Convolutional neural networks for sentence classification.
\newblock In \emph{Proceedings of the 2014 Conference on Empirical Methods in
  Natural Language Processing (EMNLP)}, 2014.

\bibitem[Kingma \& Ba(2014)Kingma and Ba]{kingma2014adam}
Kingma, Diederik and Ba, Jimmy.
\newblock Adam: A method for stochastic optimization.
\newblock \emph{arXiv preprint arXiv:1412.6980}, 2014.

\bibitem[Kiros et~al.(2015)Kiros, Zhu, Salakhutdinov, Zemel, Torralba, Urtasun,
  and Fidler]{kiros2015skip}
Kiros, Ryan, Zhu, Yukun, Salakhutdinov, Ruslan, Zemel, Richard~S, Torralba,
  Antonio, Urtasun, Raquel, and Fidler, Sanja.
\newblock Skip-thought vectors.
\newblock \emph{arXiv preprint arXiv:1506.06726}, 2015.

\bibitem[Le \& Mikolov(2014)Le and Mikolov]{le2014distributed}
Le, Quoc~V and Mikolov, Tomas.
\newblock Distributed representations of sentences and documents.
\newblock \emph{arXiv preprint arXiv:1405.4053}, 2014.

\bibitem[Li et~al.(2015{\natexlab{a}})Li, Luong, and
  Jurafsky]{li2015hierarchical}
Li, Jiwei, Luong, Minh-Thang, and Jurafsky, Dan.
\newblock A hierarchical neural autoencoder for paragraphs and documents.
\newblock \emph{arXiv preprint arXiv:1506.01057}, 2015{\natexlab{a}}.

\bibitem[Li et~al.(2015{\natexlab{b}})Li, Luong, and
  Jurafsky]{li-luong-jurafsky:2015:ACL-IJCNLP}
Li, Jiwei, Luong, Thang, and Jurafsky, Dan.
\newblock A hierarchical neural autoencoder for paragraphs and documents.
\newblock In \emph{Proceedings of the 53rd Annual Meeting of the Association
  for Computational Linguistics and the 7th International Joint Conference on
  Natural Language Processing (Volume 1: Long Papers)}, 2015{\natexlab{b}}.

\bibitem[Ling et~al.(2015)Ling, Dyer, Black, Trancoso, Fermandez, Amir, Marujo,
  and Luis]{ling-EtAl:2015:EMNLP2}
Ling, Wang, Dyer, Chris, Black, Alan~W, Trancoso, Isabel, Fermandez, Ramon,
  Amir, Silvio, Marujo, Luis, and Luis, Tiago.
\newblock Finding function in form: Compositional character models for open
  vocabulary word representation.
\newblock In \emph{Proceedings of the 2015 Conference on Empirical Methods in
  Natural Language Processing}, 2015.

\bibitem[Liu et~al.(2015)Liu, Qiu, Chen, Wu, and Huang]{liu-EtAl:2015:EMNLP2}
Liu, Pengfei, Qiu, Xipeng, Chen, Xinchi, Wu, Shiyu, and Huang, Xuanjing.
\newblock Multi-timescale long short-term memory neural network for modelling
  sentences and documents.
\newblock In \emph{Proceedings of the 2015 Conference on Empirical Methods in
  Natural Language Processing}, 2015.

\bibitem[Lu et~al.(2015)Lu, Wang, Bansal, Gimpel, and
  Livescu]{lu-EtAl:2015:NAACL-HLT}
Lu, Ang, Wang, Weiran, Bansal, Mohit, Gimpel, Kevin, and Livescu, Karen.
\newblock Deep multilingual correlation for improved word embeddings.
\newblock In \emph{Proceedings of the 2015 Conference of the North American
  Chapter of the Association for Computational Linguistics: Human Language
  Technologies}, 2015.

\bibitem[Manning et~al.(2014)Manning, Surdeanu, Bauer, Finkel, Bethard, and
  McClosky]{manning-EtAl:2014:P14-5}
Manning, Christopher~D., Surdeanu, Mihai, Bauer, John, Finkel, Jenny, Bethard,
  Steven~J., and McClosky, David.
\newblock The {Stanford} {CoreNLP} natural language processing toolkit.
\newblock In \emph{Proceedings of 52nd Annual Meeting of the Association for
  Computational Linguistics: System Demonstrations}, 2014.

\bibitem[Marelli et~al.(2014)Marelli, Bentivogli, Baroni, Bernardi, Menini, and
  Zamparelli]{marelli2014semeval}
Marelli, Marco, Bentivogli, Luisa, Baroni, Marco, Bernardi, Raffaella, Menini,
  Stefano, and Zamparelli, Roberto.
\newblock {SemEval}-2014 task 1: Evaluation of compositional distributional
  semantic models on full sentences through semantic relatedness and textual
  entailment.
\newblock In \emph{Proceedings of the 8th International Workshop on Semantic
  Evaluation ({SemEval} 2014)}, 2014.

\bibitem[Mikolov et~al.(2013)Mikolov, Sutskever, Chen, Corrado, and
  Dean]{mikolov2013distributed}
Mikolov, Tomas, Sutskever, Ilya, Chen, Kai, Corrado, Greg~S, and Dean, Jeff.
\newblock Distributed representations of words and phrases and their
  compositionality.
\newblock In \emph{Advances in Neural Information Processing Systems}, 2013.

\bibitem[Mitchell \& Lapata(2008)Mitchell and Lapata]{mitchell2008vector}
Mitchell, Jeff and Lapata, Mirella.
\newblock Vector-based models of semantic composition.
\newblock In \emph{Proceedings of the 46th Annual Meeting of the Association
  for Computational Linguistics}, 2008.

\bibitem[Mitchell \& Lapata(2010)Mitchell and Lapata]{Mitchell:Lapata:2010}
Mitchell, Jeff and Lapata, Mirella.
\newblock Composition in distributional models of semantics.
\newblock \emph{Cognitive Science}, 34\penalty0 (8), 2010.

\bibitem[Paperno et~al.(2014)Paperno, Pham, and
  Baroni]{paperno-pham-baroni:2014:P14-1}
Paperno, Denis, Pham, Nghia~The, and Baroni, Marco.
\newblock A practical and linguistically-motivated approach to compositional
  distributional semantics.
\newblock In \emph{Proceedings of the 52nd Annual Meeting of the Association
  for Computational Linguistics (Volume 1: Long Papers)}, 2014.

\bibitem[Pascanu et~al.(2012)Pascanu, Mikolov, and
  Bengio]{pascanu2012difficulty}
Pascanu, Razvan, Mikolov, Tomas, and Bengio, Yoshua.
\newblock On the difficulty of training recurrent neural networks.
\newblock \emph{arXiv preprint arXiv:1211.5063}, 2012.

\bibitem[Pavlick et~al.(2015)Pavlick, Rastogi, Ganitkevich, Durme, and
  Callison-Burch]{PavlickEtAl-2015:ACL:PPDB2.0}
Pavlick, Ellie, Rastogi, Pushpendre, Ganitkevich, Juri, Durme, Benjamin~Van,
  and Callison-Burch, Chris.
\newblock {PPDB} 2.0: Better paraphrase ranking, fine-grained entailment
  relations, word embeddings, and style classification.
\newblock In \emph{Proceedings of the Annual Meeting of the Association for
  Computational Linguistics}, 2015.

\bibitem[Pennington et~al.(2014)Pennington, Socher, and
  Manning]{pennington2014glove}
Pennington, Jeffrey, Socher, Richard, and Manning, Christopher~D.
\newblock Glove: Global vectors for word representation.
\newblock \emph{Proceedings of Empirical Methods in Natural Language Processing
  (EMNLP 2014)}, 2014.

\bibitem[Pham et~al.(2015)Pham, Kruszewski, Lazaridou, and
  Baroni]{pham-EtAl:2015:ACL-IJCNLP}
Pham, Nghia~The, Kruszewski, Germ\'{a}n, Lazaridou, Angeliki, and Baroni,
  Marco.
\newblock Jointly optimizing word representations for lexical and sentential
  tasks with the c-phrase model.
\newblock In \emph{Proceedings of the 53rd Annual Meeting of the Association
  for Computational Linguistics and the 7th International Joint Conference on
  Natural Language Processing (Volume 1: Long Papers)}, 2015.

\bibitem[Polajnar et~al.(2015)Polajnar, Rimell, and
  Clark]{polajnar-rimell-clark:2015:LSDSem}
Polajnar, Tamara, Rimell, Laura, and Clark, Stephen.
\newblock An exploration of discourse-based sentence spaces for compositional
  distributional semantics.
\newblock In \emph{Proceedings of the First Workshop on Linking Computational
  Models of Lexical, Sentential and Discourse-level Semantics}, 2015.

\bibitem[Socher et~al.(2011)Socher, Huang, Pennin, Manning, and
  Ng]{SocherEtAl2011:PoolRAE}
Socher, Richard, Huang, Eric~H, Pennin, Jeffrey, Manning, Christopher~D, and
  Ng, Andrew~Y.
\newblock Dynamic pooling and unfolding recursive autoencoders for paraphrase
  detection.
\newblock In \emph{Advances in Neural Information Processing Systems}, 2011.

\bibitem[Socher et~al.(2012)Socher, Huval, Manning, and
  Ng]{socher-EtAl:2012:EMNLP-CoNLL}
Socher, Richard, Huval, Brody, Manning, Christopher~D., and Ng, Andrew~Y.
\newblock Semantic compositionality through recursive matrix-vector spaces.
\newblock In \emph{Proceedings of the 2012 Joint Conference on Empirical
  Methods in Natural Language Processing and Computational Natural Language
  Learning}, 2012.

\bibitem[Socher et~al.(2013)Socher, Perelygin, Wu, Chuang, Manning, Ng, and
  Potts]{socher-13}
Socher, Richard, Perelygin, Alex, Wu, Jean, Chuang, Jason, Manning,
  Christopher~D., Ng, Andrew, and Potts, Christopher.
\newblock Recursive deep models for semantic compositionality over a sentiment
  treebank.
\newblock In \emph{Proceedings of the 2013 Conference on Empirical Methods in
  Natural Language Processing}, 2013.

\bibitem[Socher et~al.(2014)Socher, Karpathy, Le, Manning, and
  Ng]{SocherKLMN14}
Socher, Richard, Karpathy, Andrej, Le, Quoc~V., Manning, Christopher~D., and
  Ng, Andrew~Y.
\newblock Grounded compositional semantics for finding and describing images
  with sentences.
\newblock \emph{{TACL}}, 2, 2014.

\bibitem[Sutskever et~al.(2014)Sutskever, Vinyals, and
  Le]{sutskever2014sequence}
Sutskever, Ilya, Vinyals, Oriol, and Le, Quoc~VV.
\newblock Sequence to sequence learning with neural networks.
\newblock In \emph{Advances in Neural Information Processing Systems}, 2014.

\bibitem[Tai et~al.(2015)Tai, Socher, and Manning]{tai2015improved}
Tai, Kai~Sheng, Socher, Richard, and Manning, Christopher~D.
\newblock Improved semantic representations from tree-structured long
  short-term memory networks.
\newblock \emph{arXiv preprint arXiv:1503.00075}, 2015.

\bibitem[Tian et~al.(2015)Tian, Okazaki, and Inui]{DBLP:journals/corr/TianOI15}
Tian, Ran, Okazaki, Naoaki, and Inui, Kentaro.
\newblock The mechanism of additive composition.
\newblock \emph{arXiv preprint arXiv:1511.08407}, 2015.

\bibitem[Turian et~al.(2010)Turian, Ratinov, and
  Bengio]{Turian10wordrepresentations}
Turian, Joseph, Ratinov, Lev-Arie, and Bengio, Yoshua.
\newblock Word representations: A simple and general method for semi-supervised
  learning.
\newblock In \emph{Proceedings of the 48th Annual Meeting of the Association
  for Computational Linguistics}, 2010.

\bibitem[Vinyals et~al.(2014)Vinyals, Kaiser, Koo, Petrov, Sutskever, and
  Hinton]{vinyals2014grammar}
Vinyals, Oriol, Kaiser, Lukasz, Koo, Terry, Petrov, Slav, Sutskever, Ilya, and
  Hinton, Geoffrey.
\newblock Grammar as a foreign language.
\newblock \emph{arXiv preprint arXiv:1412.7449}, 2014.

\bibitem[Wang \& Nyberg(2015)Wang and Nyberg]{wang-nyberg:2015:ACL-IJCNLP}
Wang, Di and Nyberg, Eric.
\newblock A long short-term memory model for answer sentence selection in
  question answering.
\newblock In \emph{Proceedings of the 53rd Annual Meeting of the Association
  for Computational Linguistics and the 7th International Joint Conference on
  Natural Language Processing (Volume 2: Short Papers)}, 2015.

\bibitem[Wen et~al.(2015)Wen, Gasic, Mrk\v{s}i\'{c}, Su, Vandyke, and
  Young]{wen-EtAl:2015:EMNLP}
Wen, Tsung-Hsien, Gasic, Milica, Mrk\v{s}i\'{c}, Nikola, Su, Pei-Hao, Vandyke,
  David, and Young, Steve.
\newblock Semantically conditioned {LSTM}-based natural language generation for
  spoken dialogue systems.
\newblock In \emph{Proceedings of the 2015 Conference on Empirical Methods in
  Natural Language Processing}, 2015.

\bibitem[Weston et~al.(2010)Weston, Bengio, and Usunier]{weston2010large}
Weston, Jason, Bengio, Samy, and Usunier, Nicolas.
\newblock Large scale image annotation: learning to rank with joint word-image
  embeddings.
\newblock \emph{Machine learning}, 81\penalty0 (1), 2010.

\bibitem[Wieting et~al.(2015)Wieting, Bansal, Gimpel, Livescu, and
  Roth]{wieting2015ppdb}
Wieting, John, Bansal, Mohit, Gimpel, Kevin, Livescu, Karen, and Roth, Dan.
\newblock From paraphrase database to compositional paraphrase model and back.
\newblock \emph{Transactions of the Association for Computational Linguistics},
  3, 2015.

\bibitem[Xu et~al.(2015{\natexlab{a}})Xu, Ba, Kiros, Cho, Courville,
  Salakhutdinov, Zemel, and Bengio]{DBLP:conf/icml/XuBKCCSZB15}
Xu, Kelvin, Ba, Jimmy, Kiros, Ryan, Cho, Kyunghyun, Courville, Aaron~C.,
  Salakhutdinov, Ruslan, Zemel, Richard~S., and Bengio, Yoshua.
\newblock Show, attend and tell: Neural image caption generation with visual
  attention.
\newblock In \emph{Proceedings of the 32nd International Conference on Machine
  Learning, {ICML}}, 2015{\natexlab{a}}.

\bibitem[Xu et~al.(2015{\natexlab{b}})Xu, Callison-Burch, and
  Dolan]{xu2015semeval}
Xu, Wei, Callison-Burch, Chris, and Dolan, William~B.
\newblock {SemEval}-2015 task 1: Paraphrase and semantic similarity in
  {Twitter} ({PIT}).
\newblock In \emph{Proceedings of the 9th International Workshop on Semantic
  Evaluation ({SemEval})}, 2015{\natexlab{b}}.

\bibitem[Xu et~al.(2015{\natexlab{c}})Xu, Mou, Li, Chen, Peng, and
  Jin]{xu-EtAl:2015:EMNLP2}
Xu, Yan, Mou, Lili, Li, Ge, Chen, Yunchuan, Peng, Hao, and Jin, Zhi.
\newblock Classifying relations via long short term memory networks along
  shortest dependency paths.
\newblock In \emph{Proceedings of the 2015 Conference on Empirical Methods in
  Natural Language Processing}, 2015{\natexlab{c}}.

\bibitem[Yih et~al.(2011)Yih, Toutanova, Platt, and Meek]{yih-EtAl:2011:CoNLL}
Yih, Wen-tau, Toutanova, Kristina, Platt, John~C., and Meek, Christopher.
\newblock Learning discriminative projections for text similarity measures.
\newblock In \emph{Proceedings of the Fifteenth Conference on Computational
  Natural Language Learning}, 2011.

\bibitem[Yin \& Sch\"{u}tze(2015)Yin and
  Sch\"{u}tze]{yin-schutze:2015:NAACL-HLT1}
Yin, Wenpeng and Sch\"{u}tze, Hinrich.
\newblock Convolutional neural network for paraphrase identification.
\newblock In \emph{Proceedings of the 2015 Conference of the North American
  Chapter of the Association for Computational Linguistics: Human Language
  Technologies}, 2015.

\bibitem[Yu \& Dredze(2015)Yu and Dredze]{TACL586}
Yu, Mo and Dredze, Mark.
\newblock Learning composition models for phrase embeddings.
\newblock \emph{Transactions of the Association for Computational Linguistics},
  3, 2015.

\bibitem[Zhao et~al.(2015)Zhao, Lu, and Poupart]{zhao2015self}
Zhao, Han, Lu, Zhengdong, and Poupart, Pascal.
\newblock Self-adaptive hierarchical sentence model.
\newblock In \emph{Proceedings of IJCAI}, 2015.

\end{thebibliography}
\bibliographystyle{iclr2016_conference}

\end{document}